\documentclass[11pt,a4paper]{article}
\usepackage[hyperref]{acl2017}
\usepackage{times}
\usepackage{latexsym}

\usepackage{url}

\usepackage{graphicx}
\usepackage{amssymb}
\usepackage{amsmath}
\usepackage{xspace}
\usepackage{subfig}
\aclfinalcopy 

\newcommand{\eat}[1]{}

\newcommand{\ie}{\emph{i.e.,}\xspace}

\newcommand{\etal}{\emph{et al.}\xspace}

\newcommand{\hide}[1]{\iffalse #1 \fi}
\newcommand{\explain}[1]{\textcolor{red}{#1}}

\title{Character-Based Text Classification using Top Down Semantic Model for Sentence Representation} 

\author{Zhenzhou Wu$^1$ \\
  {\tt hyciswu@gmail.com} \And
  Xin Zheng$^{1,2}$ \\
  {\tt xin.zheng@sap.com} \And
  Daniel Dahlmeier$^1$ \\
  {\tt d.dahlmeier@sap.com} \AND\\
  SAP Innovation Center Singapore$^1$ \\
  Nanyang Technological University, Singapore$^2$ \\
  }

\begin{document}

\maketitle

\begin{abstract}

Despite the success of deep learning on many fronts especially image and speech, its application in text classification often is still not as good as a simple linear SVM on n-gram TF-IDF representation especially for smaller datasets. Deep learning tends to emphasize on sentence level semantics when learning a representation with models like recurrent neural network or recursive neural network, however from the success of TF-IDF representation, it seems a bag-of-words type of representation has its strength. Taking advantage of both representions, we present a model known as TDSM (Top Down Semantic Model) for extracting a sentence representation that considers both the word-level semantics by linearly combining the words with attention weights and the sentence-level semantics with BiLSTM and use it on text classification. We apply the model on characters and our results show that our model is better than all the other character-based and word-based convolutional neural network models by \cite{zhang15} across seven different datasets with only 1\% of their parameters. We also demonstrate that this model beats traditional linear models on TF-IDF vectors on small and polished datasets like news article in which typically deep learning models surrender.
\end{abstract}

\section{Introduction}


Recently, deep learning has been particularly successful in speech and image as an automatic feature extractor \cite{alex2013a, olga2014, km2015}, however deep learning's application to text as an automatic feature extractor has not been always successful \cite{zhang15} even compared to simple linear models with BoW or TF-IDF feature representation. In many experiments when the text is polished like news articles or when the dataset is small, BoW or TF-IDF is still the state-of-art representation compared to sent2vec or paragraph2vec \cite{quoc14} representation using deep learning models like RNN (Recurrent Neural Network) or CNN (Convolution Neural Network) \cite{zhang15}. It is only when the dataset becomes large or when the words are noisy and non-standardized with misspellings, text emoticons and short-forms that deep learning models which learns the sentence-level semantics start to outperform BoW representation, because under such circumstances, BoW representation can become extremely sparse and the vocabulary size can become huge. It becomes clear that for large, complex data, a large deep learning model with a large capacity can extract a better sentence-level representation than BoW sentence representation. However, for small and standardized news-like dataset, a direct word counting TF-IDF sentence representation is superior. Then the question is can we design a deep learning model that performs well for both simple and complex, small and large datasets? And when the dataset is small and standardized, the deep learning model should perform comparatively well as BoW? With that problem in mind, we designed TDSM (Top-Down-Semantic-Model) which learns a sentence representation that carries the information of both the BoW-like representation and RNN style of sentence-level semantic which performs well for both simple and complex, small and large datasets.

Getting inspiration from the success of TF-IDF representation, our model intends to learn a word topic-vector which is similar to TF-IDF vector of a word but is different from word embedding, whereby the values in the topic-vector are all positives, and each dimension of the topic-vector represents a topic aspect of the word. Imagine a topic-vector of representation meaning $[animal, temperature, speed]$, so a $rat$ maybe represented as $[0.9, 0.7, 0.2]$ since $rat$ is an animal with high body temperature but slow running speed compared to a $car$ which maybe represented as $[0.1, 0.8, 0.9]$ for being a non-animal, but high engine temperature and fast speed. A topic-vector will have a much richer semantic meaning than one-hot TF-IDF representation and also, it does not have the cancellation effect of summing word-embeddings positional vectors $([-1, 1] + [1, -1] = [0, 0])$. The results from \cite{johnson2015semi} show that by summing word-embedding vectors as sentence representation will have a catastrophic result for text classification.

Knowing the topic-vector of each word, we can combine the words into a sentence representation $\tilde{s}$ by learning a weight $w_i$ for each word ${v_i}$ and do a linear sum of the words, $\tilde{s} = \sum_i {w_i}\tilde{v_i}$. The weights $w_i$ for each word in the sentence summation is learnt by recurrent neural network (RNN) \cite{razvan13} with attention over the words \cite{yang2016hierarchical}. The weights corresponds to the IDF (inverse document frequency) in TF-IDF representation, but with more flexibility and power. IDF is fixed for each word and calculated from all the documents (entire dataset), however attention weights learned from RNN is conditioned on both the document-level and dataset-level semantics. This sentence representation from topic-vector of each word is then concatenated with the sentence-level semantic vector from RNN to give a top-down sentence representation as illustrated in Figure \ref{f1:model_illustration}.

\begin{figure*}[t]
	\vspace{-0.2in}
	\centering
	\includegraphics[width=0.4\textwidth]{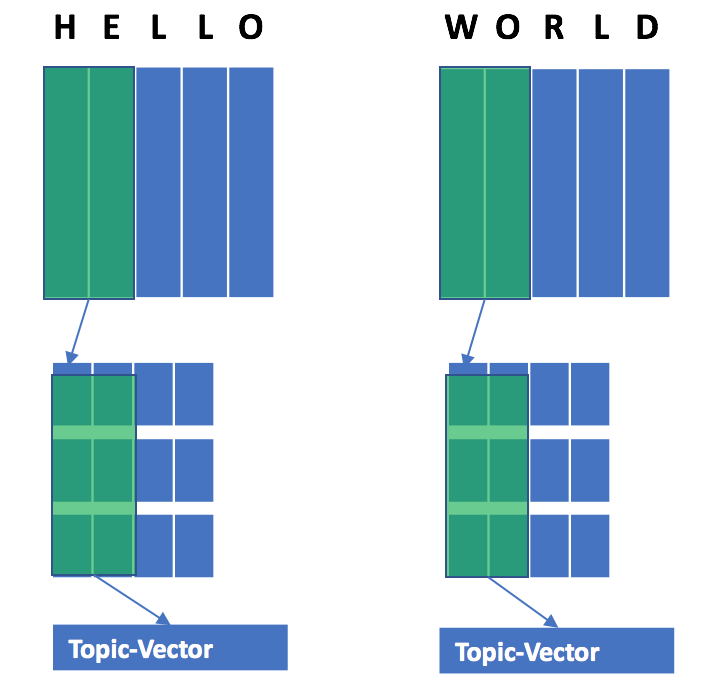}
	\vspace{-0.2in}
	\caption{Illustration of transformation from character-level embeddings to word-level topic-vector. The transformation is done with fully convolutional network (FCN) similar to \cite{long2015fully}, each hierarchical level of the FCN will extract an n-gram character feature of the word until the word-level topic-vector.
	}
	\label{f1:topic-vector}
\end{figure*}

\begin{figure*}[ht]
	\vspace{-0.2in}
	\centering
	\includegraphics[width=0.8\textwidth]{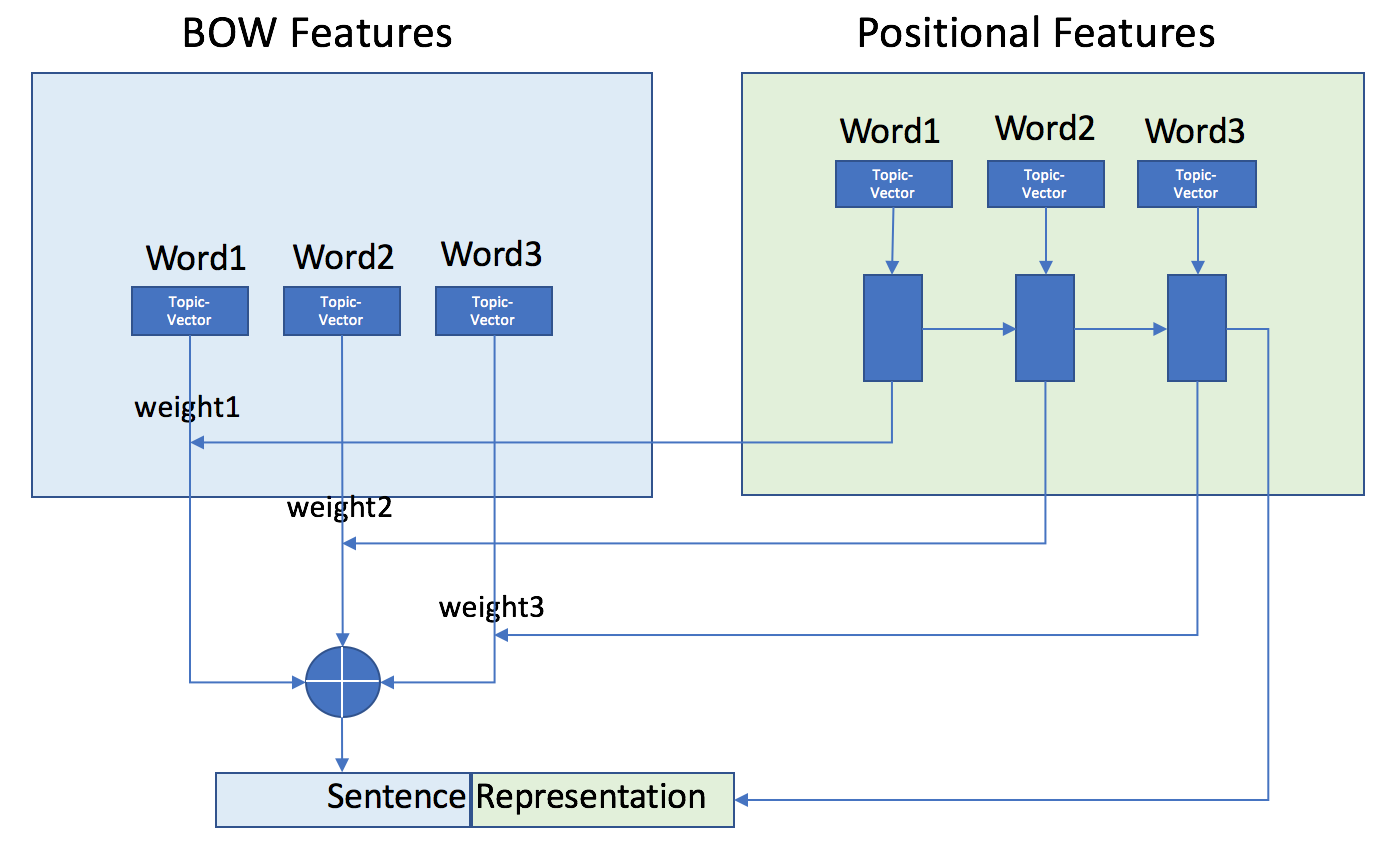}
	\vspace{-0.1in}
	\caption{TDSM: Illustration of how topic-vector of the words are combined together into a sentence representation. Note that for the actual model, we are using BiLSTM for extracting positional features. In this diagram, in order to present our idea in a neater manner, we demonstrate with a vanilla RNN.}
	\label{f1:model_illustration}
\end{figure*}

\subsection{TDSM on Characters}
TDSM is a framework that can be applied to both word-level or character-level inputs. Here in this paper, we choose character-level over word-level inputs for practical industry reasons.
\begin{enumerate}
    \item In industry applications, often the model is required to have continuous learning on datasets that morph over time. Which means the vocabulary may change over time, therefore feeding the dataset by characters dispel the need for rebuilding a new vocabulary every time when there are new words.
    \item Industry datasets are usually very complex and noisy with a large vocabulary, therefore the memory foot-print of storing word embeddings is much larger than character embeddings.
\end{enumerate}
Therefore improving the performance of a character-based model has a much larger practical value compared to word-based model.

\hide{To deal with these issues, we propose a model which could capture various kinds of features that could benefit classification task.

In this paper, we learn character representation and combine both character-level~\cite{zhang15} and word-level embedding to represent a word. Thus both morphology and semantic properties of the word are captured. As we know, not all the words in a sentence contribute equally to predicting the sentence's label. Therefore, assigning the relative word importance would give better chance of correct prediction. Attention mechanism~\cite{DBLP:conf/nips/MnihHGK14,DBLP:conf/icassp/BahdanauCSBB16} which focuses on specific part of input could help achieve this goal. The applications of attention mechanism are mostly on sequential model, while we employ the idea of attention on a feed-forward network~\cite{DBLP:journals/corr/RaffelE15}. By multiplying the weight assigned by attention mechanism to its corresponding word vector, }

\hide{a weighted feature matrix could be constructed by concatenating the sequence of word embeddings in a sentence.

\explain{Short text usually could not provide much useful information for class prediction.} We try different views to extract as much information as possible to construct an enriched sentence representation vector. Specifically, to convert a sentence representation matrix to an \explain{enriched vector}, we draw two types of features. The first one is based on word feature space and the other one is based on n-gram. However, not all the features contribute the same on sentence classification. Attention mechanism is applied to focus on the significant features. Since these features come from different views, we need a method to make the elements consistent. The residual network proposed in~\cite{km2015,DBLP:conf/eccv/HeZRS16} achieve much better results on image classification task. In other words, the residual mechanism could construct better image representation. Therefore, we adopt residual network to refine the sentence representation vector. \hide{Once we obtain a good quality representation for the sentence, it will be delivered to a classifier.}
}

\section{Related Work}

There are many traditional machine learning methods for text classification and most of them could achieve quite good results on formal text datasets. Recently, many deep learning methods have been proposed to solve the text classification task~\cite{zhang15,DBLP:conf/coling/SantosG14,DBLP:conf/emnlp/Kim14}.

Deep convolutional neural network has been extremely successful for image classification~\cite{DBLP:conf/nips/KrizhevskySH12,DBLP:journals/corr/SermanetEZMFL13}. Recently, many research also tries to apply it on text classification problem.
Kim~\citeyear{DBLP:conf/emnlp/Kim14} proposed a model similar to Collobert's \etal~\citeyear{DBLP:journals/jmlr/CollobertWBKKK11} architecture. However, they employ two channels of word vectors. One is static throughout training and the other is fine-tuned via back-propagation. Various size of filters are applied on both channels, and the outputs are concatenated together. Then max-pooling over time is taken to select the most significant feature among each filter. The selected features are concatenated as the sentence vector.

Similarly, Zhang \etal~\citeyear{zhang15} also employs the convolutional networks but on characters instead of words for text classification. They design two networks for the task, one large and one small. Both of them have nine layers including six convolutional layers and three fully-connected layers. Between the three fully connected layers they insert two dropout layers for regularization. For both convolution and max-pooling layers, they employ 1-D filters ~\cite{DBLP:conf/cvpr/BoureauBLP10}. After each convolution, they apply 1-D max-pooling. Specially, they claim that 1-D max-pooling enable them to train a relatively deep network.

Besides applying models directly on testing datasets, more aspects are considered when extracting features. Character-level feature is adopted in many tasks besides Zhang \etal~\citeyear{zhang15} and most of them achieve quite good performance.

Santos and Zadrozny~\citeyear{DBLP:conf/icml/SantosZ14} take word morphology and shape into consideration which have been ignored for part-of-speech tagging task. They suggest the intra-word information is extremely useful when dealing with morphologically rich languages. They adopt neural network model to learn the character-level representation which is further delivered to help word embedding learning.

Kim \etal~\citeyear{kim2015character} constructs neural language model by analysis of word representation obtained from character composition. Results suggest that the model could encode semantic and orthographic information from character level.

\cite{tangdocument, yang2016hierarchical} uses two hierarchies of recurrent neural network to extract the document representation. The lower hierarchical recurrent neural network summarizes a sentence representation from the words in the sentence. The upper hierarchical neural network then summarizes a document representation from the sentences in the document. The major difference between \cite{tangdocument} and \cite{yang2016hierarchical} is that Yang applies attention over outputs from the recurrent when learning a summarizing representation.

Attention model is also utilized in our model, which is used to assign weights for each word. Usually, attention is used in sequential model~\cite{DBLP:journals/corr/RocktaschelGHKB15,DBLP:conf/nips/MnihHGK14,DBLP:conf/icassp/BahdanauCSBB16,DBLP:conf/acl/KadlecSBK16}. The attention mechanism includes sensor, internal state, actions and reward. At each time-step, the sensor will capture a glimpse of the input which is a small part of the entire input. Internal state will summarize the extracted information. Actions will decide the next glimpse location for the next step and reward suggests the benefit when taking the action. In our network, we adopt a simplified attention network as~\cite{DBLP:journals/corr/RaffelE15,DBLP:conf/icassp/RaffelE16a}. We learn the weights over the words directly instead of through a sequence of actions and rewards.

Residual network~\cite{km2015,DBLP:conf/eccv/HeZRS16,DBLP:conf/ccpr/ChenCCWL16} is known to be able to make very deep neural networks by having skip-connections that allows gradient to back-propagate through the skip-connections. Residual network in~\cite{km2015} outperforms the state-of-the-art models on image recognition. He \citeyear{DBLP:conf/eccv/HeZRS16} introduces residual block as similar to feature refinement for image classification. Similarly, for text classification problem, the quality of sentence representation is also quite important for the final result. Thus, we try to adopt the residual block as in~\cite{km2015,DBLP:conf/eccv/HeZRS16} to refine the sentence vector.

\begin{table*}[t]
	\caption{Statistics of datasets.}
	\label{t1:data_statistcs}
	\begin{center}
		\begin{tabular}{ccccc}
			\hline
			{\bf Dataset}  &{\bf Classes} &{\bf Train Samples} &{\bf Test Samples} &{\bf Average Length of Text}\\
			\hline
			AG News       & 5 & 120,000   & 7,600 & 38 \\
			Sogou News    & 5 & 450,000   & 60,000& 566 \\
			DBP           & 14& 560,000   & 70,000& 48 \\
			Yelp Polarity & 2 & 560,000   & 38,000& 133 \\
			Yelp Full     & 5 & 650,000   & 50,000& 134 \\
	        Amazon Full   & 5 &3,000,000  &650,000& 80\\
			Amazon Polarity &2&3,600,000  &400,000& 78 \\

			\hline
		\end{tabular}
	\end{center}
\end{table*}

\section{Model}
\label{sec:model}
\vspace{-0.1cm}
\subsection{Characters to Topic-Vector}
Unlike word-embedding \cite{mikolov2013distributed}, topic-vector tries to learn a distributed topic representation at each dimension of the representation vector, which thus allows the simple addition of the word-level topic-vectors to form a sentence representation. Figure \ref{f1:topic-vector} illustrates how topic-vector is extracted from characters in words using FCN (Fully Convolutional Network). In order to force word level representation with topic meanings, we apply a sigmoid function over the output from FCN. Doing so, restrain the values at each dimension to be between 0 and 1, thus forcing the model to learn a distributed topic representation of the word.

\hide{
\begin{figure*}[t]
	\vspace{.3in}
	\centering
	\includegraphics[width=0.8\textwidth]{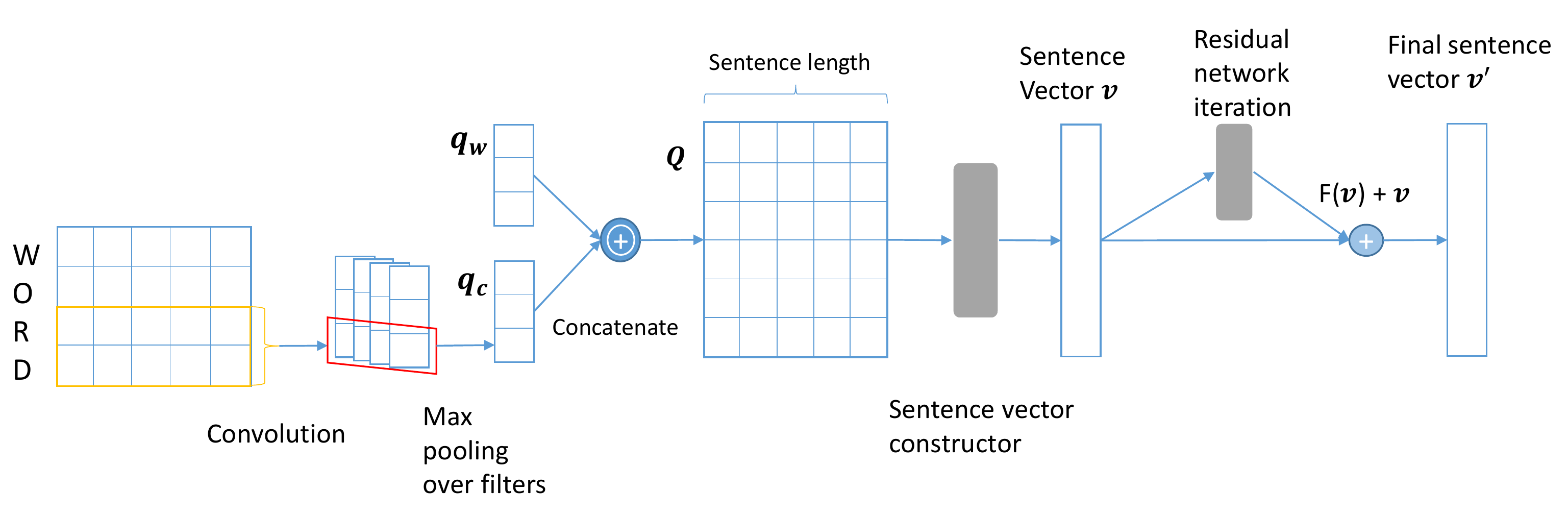}
	\vspace{.3in}
	\caption{Illustration of the proposed model. $q_{c}$ is the character-level embedding vector for the word, and $q_{w}$ is the word embedding generated according to~\cite{DBLP:conf/nips/MikolovSCCD13}. Column of $\mathbf{Q}$ is the concatenation of $q_{c}$ and $q_{w}$. The row length is sentence length. The grey box for sentence vector constructor is illustrated in Figure~\ref{f2:sentence_vector} and the grey box for Residual network iteration is illustrated in Figure~\ref{f3b:resnet}.}
	\label{f1:model_illustration}
\end{figure*}
}

\hide{
In this paper, we propose a character-aware attention residual network to generate sentence representation. Figure~\ref{f1:model_illustration} illustrates the model. For each word, the word representation vector is constructed by concatenating both character-level embedding and word semantic embedding. Thus a sentence is represented by a matrix. Then two types of features are extracted from the sentence matrix to construct the enriched sentence representation vector for short text. However, not all the features contribute the same for classification. Attention mechanism is employed to target on pertinent parts. To make features extracted from different views consistent, a residual network is adopt to refine the sentence representation vector. Thus, an enriched sentence vector is obtained to do text classification.

\subsection{Word Representation Construction}
\label{ssec:wordRep}

Let $\mathcal{C}$ be the vocabulary of characters, and $\mathbf{E} \in \mathbb{R}^{d_{c}\times |\mathcal{C}|}$ is the character embedding matrix, where $d_{c}$ is the dimensionality of character embedding. Given a word, which is composed of a sequence of characters $[c_{1}, c_{2}, ..., c_{n_{c}}]$, its corresponding character-level embedding matrix would be $ \mathbf{E}^{w} \in \mathbb{R}^{d_{c} \times n_{c}} $. Herein,
\begin{equation}
\mathbf{E}^{w}_{\cdot i} = \mathbf{E} \cdot v_{i}
\end{equation}
where $ v_{i} $ is a binary column vector with $1$ only at the $ c_{i} $-th place and $0$ for other positions. Here, we fix the word length $ d_{c} $ and take zero-padding when necessary.

For each of such matrix $\mathbf{E}^{w}$, a convolution operation~\cite{le1990handwritten} with $m$ filters (\ie kernels) $ \mathbf{P} \in \mathbb{R}^{d_{c} \times k} $ is applied on $\mathbf{E}^{w}$, and a set of feature maps could be obtained. Instead of adopting max-pooling over time~\cite{DBLP:journals/jmlr/CollobertWBKKK11}, we adopt max-pooling over filters operation to capture local information of words as shown in Figure~\ref{f1:model_illustration}. Similar operation is adopted in~\cite{DBLP:conf/cikm/ShenHGDM14}. That is we get the max feature value over results of $ m $ filters at the same window position, which depicts the most significant feature over the $ k $ characters. Thus, a vector $ \mathbf{q^{c}} $ for the word which captures the character-level information is constructed.

Note that embedding vector $ \mathbf{q^{c}} $ could only capture the word morphological features, while it can not reflect word semantic and syntactic characteristics. Therefore, we concatenate the distributed word representative vector $ \mathbf{q^{w}} $ (\ie Word2Vec)~\cite{DBLP:conf/nips/MikolovSCCD13} to $ \mathbf{q^{c}} $ as the word's final representation $ \mathbf{q} \in \mathbb{R}^{(d_{c} + d_{w})} $, where $ d_{w} $ is the dimensionality of Word2Vec. Given a sentence, which consists of a sequence of words $[w_{1}, w_{2}, ..., w_{n_{w}}]$, its representation matrix is $ \mathbf{Q} \in \mathbb{R}^{(d_{c} + d_{w}) \times n_{w}} $.
}

\subsection{Sentence Representation Vector Construction}
\label{ssec:senRep}
Forming a sentence representation from words can be done simply by summing of the word-embeddings which produce catastrophic results \cite{johnson2015semi} due to the cancellation effect of adding embedding vectors (negative plus positive gives zero). Or in our model, the summing of word-level topic-vectors which give a much better sentence representation as shown in Table \ref{t3:comparisonResults} than summing word-embeddings.

\paragraph{BoW Features:} Sentence vector derived from summing of the word topic-vectors is equivalent to the BoW vectors in word counting, whereby we treat the prior contribution of each word to the final sentence vector equally. Traditionally, a better sentence representation over BoW will be TF-IDF, which gives a weight to each word in a document in terms of IDF (inverse document frequency). Drawing inspiration from TF-IDF representation, we can have a recurrent neural network that outputs the attention \cite{yang2016hierarchical} over the words. And the attention weights serve similar function as the IDF except that it is local to the context of the document, since the attention weight for a word maybe different for different documents while the IDF of a word is the same throughout all documents. With the attention weights $w_i$ and word topic-vector $\tilde{v}_i$, we can form a sentence vector $\tilde{s}_{bow}$ by linear sum

\begin{equation}
\tilde{s}_{bow} = \sum_i w_i \tilde{v}_i
\end{equation}

\paragraph{Positional Features:} With the neural sentence vector that derived from BoW which captures the information of individual words. We can also concatenate it with the output state from the RNN which captures the document level information and whose representation is conditioned on the positioning of the words in the document.

\vspace{-0.6cm}
\begin{align}
\label{eqn:rnn}
&\tilde{s}_{t} = RNN(\tilde{v}_{t}, \tilde{s}_{t-1}) \\
&\tilde{s}_{pos} = \tilde{s}_T \\
\label{eqn:oplus}
&\tilde{s} = \tilde{s}_{bow} \oplus \tilde{s}_{pos}
\end{align}
where $T$ is the length of the document, and $\oplus$ represents concatenation such that $|\tilde{s}| = |\tilde{s}_{bow}| + |\tilde{s}_{pos}|$. The overall sentence vector $\tilde{s}$ will then capture the information of both the word-level and document-level semantics of the document. And thus it has a very rich representation.

\subsubsection{Bi-Directional LSTM}
We used Bi-directional LSTM (BiLSTM) \cite{graves2013hybrid} as the recurrent unit. BiLSTM consist of a forward LSTM (FLSTM) and a backward LSTM (BLSTM), both LSTMs are of the same design, except that FLSTM reads the sentence in a forward manner and BLSTM reads the sentence in a backward manner. One recurrent step in LSTM of Equation \ref{eqn:rnn} consists of the following steps
\begin{align}
    \tilde{f}_t &= \sigma\big(\mathbf{W}_f (\tilde{s}_{t-1} \oplus \tilde{v}_t) + \tilde{b}_f \big) \\
    \tilde{i}_t &= \sigma\big(\mathbf{W}_i (\tilde{s}_{t-1} \oplus \tilde{v}_t) + \tilde{b}_i\big) \\
    \tilde{C}_t &= \tanh\big(\mathbf{W}_C(\tilde{s}_{t-1}, \tilde{v}_t) + \tilde{b}_C\big) \\
    \tilde{C}_t &= \tilde{f}_t \otimes \tilde{C}_{t-1} + \tilde{i}_t \otimes \tilde{C}_t \\
    \tilde{o}_t &= \sigma\big(\mathbf{W}_o (\tilde{s}_{t-1} \oplus \tilde{v}_t) + \tilde{b}_o \big) \\
    \tilde{s}_t &= \tilde{o}_t \otimes \tanh(\tilde{C}_t)
\end{align}
where $\otimes$ is the element-wise vector multiplication, $\oplus$ is vector concatenation similarly defined in Equation \ref{eqn:oplus}. $\tilde{f}_t$ is forget state, $\tilde{i}_t$ is input state, $\tilde{o}_t$ is output state, $\tilde{C}_t$ is the internal context which contains the long-short term memory of historical semantics that LSTM reads. Finally, the output from the BiLSTM will be a concatenation of the output from FLSTM and BLSTM

\begin{align}
\tilde{f}_t &= \text{FLSTM}(\tilde{v}_t, \tilde{s}_{t-1}) \\
\tilde{b}_t &= \text{BLSTM}(\tilde{v}_t, \tilde{s}_{t+1}) \\
\tilde{h}_{t} &= \tilde{f}_t \oplus \tilde{b}_t
\end{align}

Here, the concatenated output state of BiLSTM has visibility of the entire sequence at any time-step compared to single-directional LSTM which only has visibility of the sequence in the past. This property of BiLSTM is very useful for learning attention weights for each word in a document because then the weights are decided based on the information of the entire document instead of just words before it as in LSTM.

\section{Experiment}
\begin{table}[th]
	\caption{Fully-Convolutional Network from characters to topic-vector. The first convolutional layer has kernel size of (100, 4) where 100 is the embedding size over 4-gram character as shown in Figure \ref{f1:topic-vector}.}
	\label{tbl:params_fcn}
	\begin{center}
		\begin{tabular}{ccccc}
			\hline
			{\bf Layer}  & {\bf No. of Filters} & {\bf Kernel Size} & {\bf Stride} \\
			\hline
			1    & 10 & (100, 4) & (1, 1)   \\
			2    & 20 & (1, 4) & (1, 1) \\
			3    & 30 & (1, 4) & (1, 2) \\
			\hline
		\end{tabular}
	\end{center}
\end{table}

\begin{table*}[t]
	\caption{Comparison results on accuracy for various models. \textit{Lg w2v Conv} and \textit{Sm. w2v Conv} is CNN on word embedding. \textit{Lg. Conv} and \textit{Sm. Conv} is CNN on character embedding. \textit{LSTM-GRNN} and \textit{HN-ATT} are different species of recurrent neural networks on words and sentences. Unfortunately, these two RNN models did not use the same text preprocessing technique as other models, so their models may not be objectively comparable to Zhang's or our model, because it is well known that \cite{zhang15, uysal2014impact}, the difference in text preprocessing will have a significant impact on the final accuracy. However, these RNN models are still a good reference for our understanding of time-based models on large datasets of long sentences.}
	\label{t3:comparisonResults}
	\begin{center}
		\begin{tabular}{l|l|l|l|l|l|l|l}
			\hline
			{\bf Method}  &{\bf AG} &{\bf Sogou} &{\bf DBP} &{\bf Yelp Po} &{\bf Yelp Full}
			&{\bf Amz Full} &{\bf Amz Po}\\
			\hline
			Dataset size (1000)    & 120 & 450 & 560 & 560 & 650 & 3000 & 3600 \\
			\hline
			\multicolumn{8}{c}{Traditional Methods} \\
			\hline
			BoW     & 88.81 & 92.85 & 96.61 & 92.24 & 57.91 & 54.64 & 90.40 \\
			TF-IDF  & 89.64 & 93.45 & 97.36 & 93.66 & 59.86 & 55.26 & 91.00 \\

			\hline
			\multicolumn{8}{c}{Word-Based Models} \\
			\hline
			Bag-of-Means   & 83.09 & 89.21 & 90.45 & 87.33 & 52.54 & 44.13 & 81.61 \\
			Lg. w2v Conv   & 90.08 & 95.61 & 98.58 & 95.40 & 59.84 & 55.60 & 94.12 \\
			Sm. w2v Conv   & 88.65 & 95.46 & 98.29 & 94.44 & 57.87 & 57.41 & 94.00 \\
			LSTM-GRNN \cite{tangdocument} & - & - & - & - & 67.6 & - & - \\
			HN-ATT \cite{yang2016hierarchical} & - & - & - & - & {\bf 71.0} & {\bf 63.6} & - \\
			\hline

			\multicolumn{8}{c}{Character-Based Models} \\
			\hline
			Lg. Conv \cite{zhang15}  & 87.18 & 95.12 & 98.27 & 94.11 & 60.38 & 58.69 & 94.49 \\
			Sm. Conv \cite{zhang15}  & 84.35 & 91.35 & 98.02 & 93.47 & 59.16 & 59.47 & 94.50 \\
            TDSM  & {\bf 90.51} & {\bf 96.17}  & {\bf 98.72} & {\bf 94.81} & {\bf 61.62}  & {\bf 60.02} & {\bf 95.08} \\
			\hline
		\end{tabular}
	\end{center}
\end{table*}

\begin{table}[t]
	\caption{Number of parameters for different models}
	\label{t1:num_params}
	\begin{center}
		\begin{tabular}{cc}
			\hline
			{\bf Model}  &{\bf No. of Parameters} \\
			\hline
			Lg. Conv \& Lg. w2v Conv        & 75,600,000   \\
			Sm. Conv \& Sm. w2v Conv      &  10,000,000\\
			TDSM    &  780,000\\

			\hline
		\end{tabular}
	\end{center}
\end{table}
\subsection{Model}
The entire model has only 780,000 parameters which is only 1\% of the parameters in \cite{zhang15} large CNN model. We used BiLSTM \cite{graves2013hybrid} with 100 units in both forward and backward LSTM cell. The output from the BiLSTM is 200 dimensions after we concatenate the outputs from the forward and backward cells. We then use attention over the words by linearly transform 200 output dimensions to 1 follow by a softmax over the 1 dimension outputs from all the words. After the characters to topic-vector transformation by FCN, each topic vector will be 180 dimensions. The topic-vectors are then linearly sum with attention weights to form a 180 dimensions BoW-like sentence vector. This vector is further concatenate with 200 dimensions BiLSTM outputs. The 380 dimensions undergo 10 blocks of ResNet \cite{DBLP:conf/ccpr/ChenCCWL16} plus one fully connected layer. We use RELU \cite{nair2010rectified} for all the intra-layer activation functions. The source code will be released after some code refactoring and we built the models with tensorflow \cite{abadi2016tensorflow} and tensorgraph \footnote{https://github.com/hycis/TensorGraph}.

\subsection{Datasets}

We use the standard benchmark datasets prepare by \cite{zhang15}. The datasets have different number of training samples and test samples ranging from 28,000 to 3,600,000 training samples, and of different text length ranging from average of 38 words for Ag News to 566 words in Sogou news as illustrated in Table \ref{t1:data_statistcs}. The datasets are a good mix of polished (AG) and noisy (Yelp and Amazon reviews), long (Sogou) and short (DBP and AG), large (Amazon reviews) and small (AG) datasets. And thus the results over these datasets serve as good evaluation on the quality of the model.





\subsection{Word Setting}

In this paper, we take $128$ ASCII characters as character set, by which most of the testing documents are composite. We define word length as $20$ and character embedding length as $100$. If a word with characters less than $20$, we will pad it with zeros. If the length is larger than $20$, we just take the first $20$ characters. We set the maximum length of words as the average number of words of the documents in the dataset plus two standard deviation, which is long enough to cover more than 97.5\% of the documents. For documents with number of words more than the preset maximum number of words, we will discard the exceeding words.

\subsection{Baseline Models}

We select both traditional models and the convolutional models from \cite{zhang15}, the recurrent models from \cite{yang2016hierarchical,tangdocument} as baselines. Also in order to ensure a fair comparison of models, such that any variation in the result is purely due to the model difference, we compare TDSM only with models that are trained in the same way of data preparation, that is the words are lowered and there are no additional data alteration or augmentation with thesaurus. Unfortunately, \cite{yang2016hierarchical, tangdocument} recurrent models are trained on full text instead of lowered text, so their models may not be objectively compared to our models, since it is well known from \cite{ uysal2014impact} that different text preprocessing will have significant impact on the final results, Zhang's result shows that a simple case lowering can result up to 4\% difference in classification accuracy. Despite this, we still include the recurrent models for comparison, because they provide a good reference for understanding time-based models on large datasets of long sentences.


\subsubsection{Traditional Models}

\paragraph{BoW} is the standard word counting method whereby the feature vector represents the term frequency of the words in a sentence.

\paragraph{TF-IDF} is similar to BoW, except that it is derived by the counting of the words in the sentence weighted by individual word's term-frequency and inverse-document-frequency \cite{joachims1998text}. This is a very competitive model especially on clean and small dataset.

\subsubsection{Word-Based Models}

\paragraph{Bag-of-Means} is derived from clustering of words-embeddings with k-means into 5000 clusters, and follow by BoW representation of the words in 5000 clusters.

\paragraph{Lg. w2v Conv and Sm w2v Conv} is CNN model on word embeddings following \cite{zhang15}, to ensure fair comparison with character-based models, the CNN architecture is the same as \textit{Lg. Conv} and \textit{Sm. Conv} with the same number of parameters.

\paragraph{LSTM-GRNN} from \cite{tangdocument} is basically a recurrent neural network based on LSTM and GRU \cite{chung2014empirical} over the words in a sentence, and over the sentences in a document. It tries to learn a hierarchical representation of the text from multi-levels of recurrent layers.

\paragraph{HN-ATT} from \cite{yang2016hierarchical} is basically similar to LSTM-GRNN except that instead of just learning the hierarchical representation of the text directly with RNN, it also learns attention weights over the words during the summarization of the words and over the sentences during the summarization of the sentences.

\subsubsection{Character-Based Models}
\paragraph{Lg. Conv and Sm. Conv} are proposed in \cite{zhang15}, which is a CNN model on character encoding and is the primary character-based baseline model that we are comparing with.

\section{Results}

Table~\ref{t3:comparisonResults} shows the comparison results of different datasets with different size, different sentence length, and different quality (polished AG news vs messy Yelp and Amazon reviews).

From the results, we see that TDSM out-performs all the other CNN models across all the datasets with only 1\% of the parameters of Zhang's large conv model and 7.8\% of his small conv model. Since these results are based on the same text preprocessing and across all kinds of dataset (long, short, large, small, polished, messy), we can confidently say that TDSM generalizes better than the other CNN models over text classification. These results show that a good architecture design can achieve a better accuracy with significantly less parameters.

Character-based models are the most significant and practical model for real large scale industry deployment because of its smaller memory footprint, agnostic to changes in vocabulary and robust to misspellings \cite{kim2015character}. For a very long time, TF-IDF has been state-of-art models especially in small and standardized datasets. However because of its large memory footprint and non-suitability for continuous learning (because a new vocabulary has to be rebuilt every once in awhile when there are new words especially for data source like Tweeter), it was not an ideal model until character-based models came out. From the results, previous character-based models are generally better than TF-IDF for large datasets but falls short for smaller dataset like AG news. TDSM successfully close the gap between character-based models and TF-IDF by beating TF-IDF with 1\% better performance. The results also confirm the hypothesis that TDSM as illustrated in Figure \ref{f1:model_illustration} which contains both the BoW-like and sentence-level features, has the best of the traditional TF-IDF and the recent deep learning model, is able to perform well for both small and large datasets.

From the results, we also observe that TDSM improves over other character-based models by a big margin of 3\% for \textit{Lg. Conv} and 5.7\% for \textit{Sm. Conv} on the AG dataset. But the improvement tails off to only 0.5\% for Amazon reviews when the dataset size increases from 120,000 to 3.6 million. This is probably because TDSM has reached its maximum capacity when the dataset gets very large compared to other character-based models which have 100 times the capacity of TDSM.

For Yelp Full, we observe that the hierarchical recurrent models \textit{LSTM-GRNN} and \textit{HN-ATT} performs about 10\% points better than TDSM but drops to only 3\% for Amazon Full. This may be partly due to their data being prepared differently from our models. This can also be due to the structure of these hierarchical recurrent models which has two levels of recurrent neural networks for summarizing a document, whereby the first level summarizes a sentence vector from the words and the second level summarizes a document vector from the sentences. So these models will start to perform much better when there are alot of sentences and words in a document. For Yelp Full, there are of average 134 words in one document and Amazon Full has about 80 words per document. That's why the performance is much better for these recurrent models on Yelp than on Amazon. However, these hierarchical recurrent models will be reduce to a purely vanilla RNN for short text like AG News or Tweets with a few sentences, and under such circumstances its result will not be much different from a standard RNN. Nevertheless, \textit{LSTM-GRNN} or \textit{HN-ATT} does indicate the strength of RNN models in summarizing the sentences and documents and deriving a coherent sentence-level and document-level representation.

\section{Conclusion}

From the results, we see a strong promise for TDSM as a competitive model for text classification because of its hybrid architecture that looks at the sentence from both the traditional TF-IDF point of view and the recent deep learning point of view. The results show that this type of view can derive a rich text representation for both small and large datasets.


\bibliography{reference}
\bibliographystyle{reference}

\end{document}